\documentclass[11pt]{article}

\usepackage[final]{acl}

\usepackage{times}
\usepackage{latexsym}

\usepackage[T1]{fontenc}

\usepackage[utf8]{inputenc}

\usepackage{microtype}

\usepackage{inconsolata}

\usepackage{graphicx}
\usepackage{booktabs}
\usepackage{multirow}
\usepackage{tabularx}
\usepackage{graphicx}
\usepackage{url}                       
\usepackage{hyperref}       
\usepackage{amsmath} 
\usepackage{bibentry}
\usepackage{url} 
\usepackage{xcolor}
\usepackage{comment}
\usepackage{amsmath,amssymb}

\usepackage{algorithm}      
\usepackage{algpseudocode}
\usepackage{mathtools}
\usepackage[table]{xcolor}
\usepackage{graphicx}
\usepackage{subcaption} 
\usepackage{amssymb}  
\usepackage{pifont}   
\usepackage{listings}        
\usepackage{xcolor}          
\usepackage{amssymb}         
\usepackage{pifont}          

\usepackage{listings}
\usepackage{xcolor}
\usepackage{caption}
\usepackage[normalem]{ulem}

\lstdefinestyle{borderedcode}{
    language=Python,
    basicstyle=\ttfamily\footnotesize,
    keywordstyle=\color{blue},
    commentstyle=\color{green},
    stringstyle=\color{red},
    showstringspaces=false,
    breaklines=true,
    frame=single,
    escapechar=!,
    numbers=none,
    backgroundcolor=\color{white}
}

\title{SrDetection: A Self-Referential Framework for Data Leakage Detection in Code Large Language Models}

\author{
 \textbf{Shuaimin Li\textsuperscript{1}},
 \textbf{Liyang Fan\textsuperscript{2,6,1}},
 \textbf{Zeyang Li\textsuperscript{3}},
 \textbf{Zhuoyue Wan\textsuperscript{4}},
 \textbf{Yufang Lin\textsuperscript{5}},
 \textbf{Shiwen Ni$^*$\textsuperscript{6}},
 \textbf{Feiteng Fang\textsuperscript{1}},\\
 \textbf{Hamid Alinejad-Rokny\textsuperscript{7}},
 \textbf{Yuanfeng Song},
 \textbf{Kun Jing\textsuperscript{8}},
 \textbf{Chen Jason Zhang\textsuperscript{4}},
 \textbf{Min Yang\thanks{Corresponding authors.}\textsuperscript{1,6}}\\
\\
 \textsuperscript{1}Shenzhen Key Laboratory for High Performance Data Mining, Shenzhen Institutes of Advanced Technology,\\ Chinese Academy of Sciences
 \textsuperscript{2}Shenzhen University
 \textsuperscript{3} University of Science and Technology of China
 \textsuperscript{4}PolyU\\
 \textsuperscript{5}East China Normal University
  \textsuperscript{6}Artificial Intelligence Research Institute, Shenzhen University of \\Advanced Technology
\textsuperscript{7}University of New South Wales
\textsuperscript{8}Anhui University\\
 \small{
   \textbf{Correspondence:} \href{mailto:email@domain}{\{sm.li2, min.yang\}@siat.ac.cn; nishiwen@suat-sz.edu.cn}
 }
}

\begin{document}
\maketitle
\begin{abstract}
Evaluating code large language models (Code LLMs) requires reliable detection of data leakage, where benchmark performance is artificially inflated by exposure to benchmark data during pre-training. Existing approaches either assume access to proprietary training corpora, rely on brittle heuristics such as timestamp filtering, or use external reference sets with manually tuned, non-generalizable thresholds. To address these limitations, we introduce \textbf{SrDetection}, a unified \textbf{s}elf-\textbf{r}eferential leakage detection framework for both gray-box (access to model logits) and black-box (access to model outputs) settings. SrDetection generates semantically equivalent variants of a benchmark sample and detects leakage by contrasting the model's behavior on the original versus its variants, flagging cases where the original is disproportionately easier for the model.
We further design a controlled leakage detection testbed and evaluate SrDetection in this environment. Across different models and training stages, SrDetection improves average F1 by 21.52 points in the gray-box setting and 14.46 points in the black-box setting over strong baselines, demonstrating robust, threshold-independent leakage detection. Finally, a gray-box study of 15 widely used Code LLMs on four popular benchmarks reveals benchmark-specific leakage patterns beyond prior overlap-based analyses\footnote{\footnotesize Source code and data are available at \url{https://github.com/SMinL/SrDetectionCode}}.
\end{abstract}

\section{Introduction}

Code Large Language Models (Code LLMs) now underpin code generation, completion, and understanding in modern software development~\cite{Jiang25codegeneration,2024arXiv241003981J,HouZLYWLLLGW24,YuSRZZMLLWX24}. As these models are increasingly evaluated and ranked by public code benchmarks, ensuring the fairness and reliability of evaluation has become critical~\cite{Zhou24icse,LopezCSSV25}. A central threat is \emph{data leakage}: if a model has already seen benchmark samples during pre-training, apparent performance gains may reflect memorization rather than generalization, inflating metrics and misleading model comparisons~\cite{MattonSATAHMVGG24,RiddellNC24}. 

This issue has two major consequences. First, it complicates the assessment of whether strong LLM performance stems from true innovation or prior exposure to benchmark data. Second, it introduces unfairness when comparing LLM-based methods to non-learning-based approaches, such as traditional program analysis, which do not rely on training data and cannot benefit from leaked samples. These concerns motivate \emph{code data leakage detection}: given a model and a single code sample, decide whether the sample was likely present in the model's pre-training corpus~\cite{LessLeak-Bench}.

Current detectors leave a key gap: they either require information that is unavailable in practice or rely on brittle assumptions that break for code. Overlap-based methods (e.g., n-gram matching) depend on access to pre-training corpora~\citep{LessLeak-Bench,LopezCSSV25}, yet leading models’ training data are typically proprietary and undisclosed~\cite{team2025chatgpt,llama3}. Time-based heuristics assume post-release code cannot be in training~\citep{LiveCodeBench}, but reuse, inheritance, and forking routinely violate this assumption. Confidence-based approaches such as Perplexity (PPL)~\citep{Benchmarking_Benchmark_Leakage,ShiAXHLB0Z24,DongJLJGYL24,Deng0TGC24,Zhang0GRFC24,abs-2311-04850} avoid corpus access, but require an external reference set to tune a threshold; these reference sets often rely on the same temporal heuristics, and thresholds calibrated on natural language transfer poorly to the structural regularities of code.

We address this gap with \textbf{SrDetection}, a \textbf{s}elf-\textbf{r}eferential leakage detector that requires no external reference data and no manually defined thresholds. The key idea is to compare a snippet to \emph{semantics-preserving variants} derived from the sample itself: a leaked sample is often unusually easy for the model in its exact surface form. We generate functionality-preserving transformations using an auxiliary LLM, forming a contrastive set. In a gray-box setting (access to logits), we compute PPL for the original and variants and flag cases where the original is consistently much easier than its variants. In a black-box setting (access only to outputs), we prompt completions from prefixes and compare surface-level similarity to true suffixes, flagging the same discrepancy pattern. By relying on \emph{relative} differences within the self-generated set, SrDetection avoids external anchors and brittle global thresholds.
To evaluate SrDetection, we construct a general testbed capable of supporting different datasets and open-source Code LLMs. We explicitly control sample inclusion through continued pre-training, yielding a reliable split between training-seen and held-out samples. This design enables precise leakage measurement under both gray-box and black-box detection scenarios, without relying on temporal heuristics.
Extensive experiments on this testbed demonstrate that SrDetection consistently outperforms strong baselines across models and training stages, achieving average F1 gains of 21.52 and 14.46 points in gray-box and black-box settings, respectively.


In summary, our main contributions are:

(1) We present \textbf{SrDetection}, a \textbf{self-referential} framework for \emph{sample-level} leakage detection that replaces external overlap checks and threshold calibration with \emph{within-sample contrast} between a code snippet and its semantics-preserving variants, covering both gray-box and black-box access.

(2) We propose an \textbf{automatic semantics-preserving code augmentation} method tailored to leakage detection, producing contrastive variants that expose memorization under common code edits and refactorings.

(3) We introduce a \textbf{controlled training-based testbed} and a \textbf{large-scale measurement study}, demonstrating robust improvements over competitive baselines and revealing benchmark-specific leakage patterns across widely used Code LLMs.
\section{Related Work}
\textbf{Data Leakage Detection for General LLMs.} Detecting whether a data sample was included in a model’s training corpus, a problem commonly studied through Membership Inference Attacks (MIAs), has a long history in machine learning~\cite{MIA,SablayrollesDSO19,SongS19}.For LLMs, many approaches leverage signals derived from output likelihoods. Perplexity (PPL) was introduced as an early baseline for membership inference~\cite{CarliniTWJHLRBS21}, followed by Min-K\% Prob~\cite{ShiAXHLB0Z24}, which averages the probabilities of the least confident tokens to obtain a more robust score. Subsequent variants refine this direction through theoretical interpretation~\cite{zhang2025mink} or calibration techniques that reduce the influence of frequent tokens~\cite{Zhang0GRFC24}.
Several works also target black-box settings where only generated text is observable. VeilProbe~\cite{VeilProbe} trains a sequence-to-sequence model to capture discrepancies between inputs and LLM outputs using token perturbations, while DPDLLM~\cite{zhou-etal-2024-dpdllm} extracts features from generated text via a reference language model for downstream membership classification. Despite methodological differences, most of these methods ultimately rely on an external reference set (or a proxy) to calibrate a decision threshold, which is commonly constructed via temporal splits or heuristic curation.

\noindent\textbf{Data Leakage Detection for Code LLMs.} 
Extending data leakage analysis to Code LLMs has recently attracted growing attention. Empirical studies reveal that large code models can memorize substantial portions of their training data verbatim~\citep{Zhou24icse}. Beyond memorization within a single dataset, inter-dataset code duplication further threatens evaluation validity~\citep{LopezCSSV25}. Several works systematically analyze contamination in code generation benchmarks, quantifying both surface-level and semantic overlap with pre-training data~\citep{RiddellNC24,MattonSATAHMVGG24,LessLeak-Bench}.
Despite these insights, most existing studies focus on empirical analysis or dataset curation, rather than providing practical detection methods. Moreover, the external reference-based strategies commonly adopted in general-domain leakage detection, which rely on timestamp-based data partitioning, are not suited for Code LLMs due to pervasive code reuse and the weak correspondence between repository timestamps and code originality.


\begin{figure*}[t]
\centering
  \includegraphics[width=\linewidth]{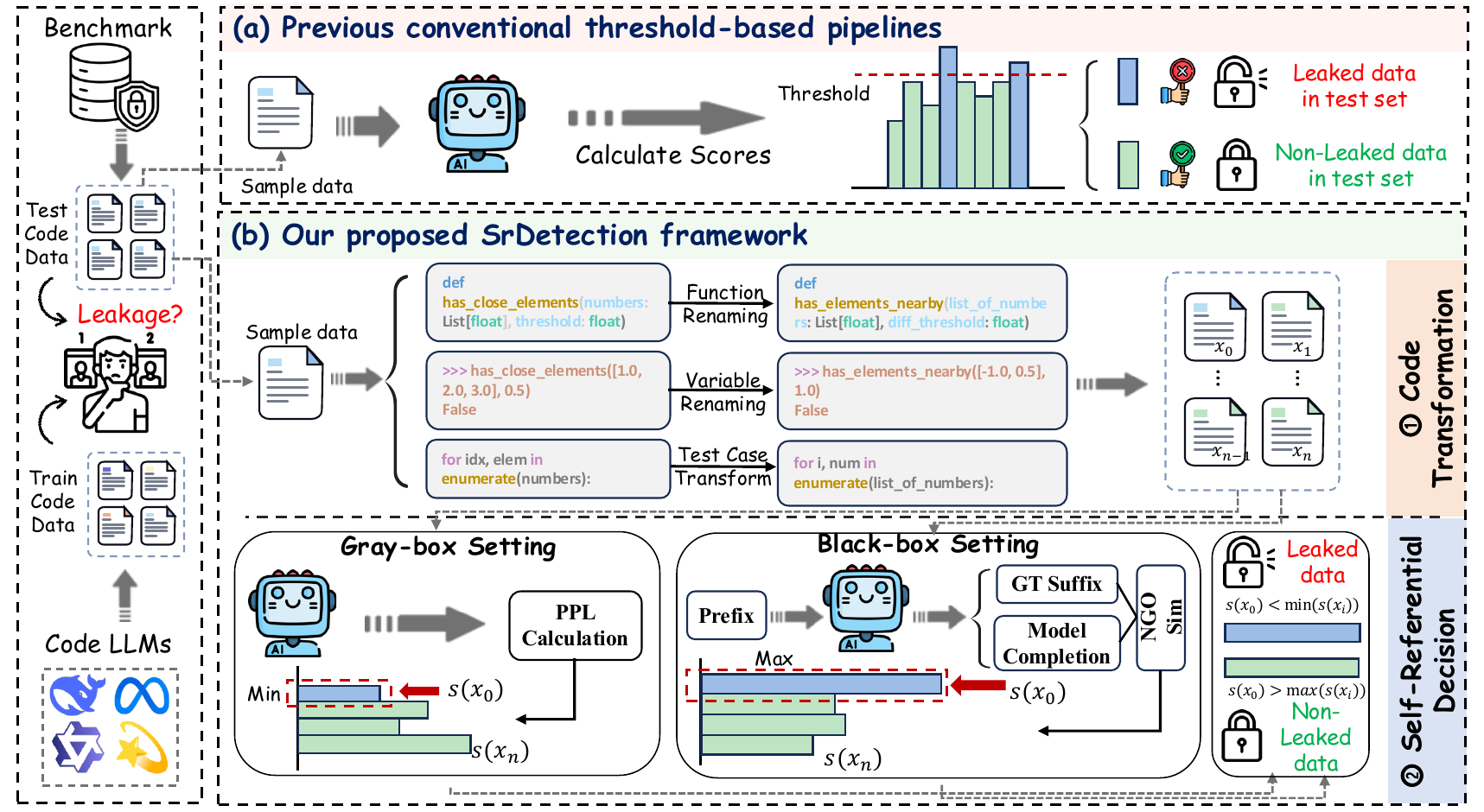} 
  \caption {Comparison between conventional threshold-based pipelines (a) and the SrDetection framework (b). SrDetection consists of a code transformation module followed by a self-referential detection module that can operate under gray-box or black-box settings to detect potential data leakage in the input sample $x$.}
  \label{fig:srdetection_framework}
\end{figure*}

\section{Methodology}
In this section, we first introduce the proposed {SrDetection}, then describe the construction of our {Leakage Testbed}, which provides a controlled testbed for assessing detection performance.

\subsection{SrDetection}
Let \( M \) denote a Code LLM and \( x \) be a sample from a benchmark dataset \( B \). If \( x \) was present in the pre-training corpus of \( M \), we consider it a \emph{leaked} sample. Our objective is to assess the degree of leakage in \( B \) with respect to \( M \). For each \( x \in B \), we aim to determine whether \( x \) has been memorized by \( M \) without relying on any external reference data or pre-defined thresholds.

We introduce SrDetection, a unified self-referential detection framework designed for both gray-box (with access to model logits) and black-box (with access only to model outputs) scenarios, as illustrated in Figure~\ref{fig:srdetection_framework}.
The core hypothesis is that if \( M \) has memorized \( x \) during pre-training, it will exhibit significantly more predictable behavior for the original form of \( x \) compared to its \textbf{semantically equivalent variants}. 
Formally, for the original sample \( x_0 \), we generate a set of \( n \) variants \( \mathcal{V} = \{x_1, x_2, \dots, x_n\} \) via semantics-preserving transformations. Detection is then performed by comparing a detection score \( \mathcal{S}_M(\cdot) \) across the set \( \mathcal{X} = \{x_0\} \cup \mathcal{V} \):
\begin{equation}
    \text{Leakage}(x) = \mathbb{I}\left[ \begin{array}{c} \text{$\mathcal{S}_M(x_0)$ is the most indicative} \\ \text{of memorization in $\mathcal{X}$} \end{array} \right],
\end{equation}
where \( \mathbb{I} \) is the indicator function, and the definition of ``most indicative'' differs between gray-box and black-box settings as detailed in subsequent sections.



Our framework consists of two main components: (1) \textbf{Code Transformation Module}, which generates semantically equivalent variants; (2) \textbf{Leakage Decision Module}, which defines the detection function $\mathbb{I}$ and detection score \( \mathcal{S}_M(\cdot) \) for leakage decision. Algorithm~\ref{alg:leakage_detection} summarizes the overall pipeline, providing a high-level view of the detection process. As illustrated in the algorithm, our method operates in three concise steps: generating semantically equivalent variants, computing detection scores under gray and black-box settings, and making a self-referential leakage decision based on relative comparison within the generated set.

\begin{algorithm}[t]
\caption{Self-Referential Code Data Leakage Detection}
\label{alg:leakage_detection}
{\footnotesize
\begin{algorithmic}[1]
\Require Code LLMs $M$, code sample $x_0$, number of variants $n$, 
access type $\mathcal{A}\in\{\mathrm{gray\text{-}box},\mathrm{black\text{-}box}\}$
\Ensure Leakage label $l\in\{0,1\}$

\Statex \Comment{Generate semantic variants}
\State $\mathcal{X}\gets\{x_0\}\cup\{x_i\}_{i=1}^n$

\Statex \Comment{Compute detection scores}
\For{each $x\in\mathcal{X}$}
\State $s(x)\gets
\begin{cases}
\mathrm{PPL}(x;M), & \mathcal{A}=\mathrm{gray\text{-}box}\\
\mathrm{NGO}(x;M), & \mathcal{A}=\mathrm{black\text{-}box}
\end{cases}$
\EndFor

\Statex \Comment{Self-referential decision}
\State $l \gets \mathbb{I}\!\left[
s(x_0)=
\begin{cases}
\min_i s(x_i), & \mathcal{A}=\mathrm{gray\text{-}box}\\
\max_i s(x_i), & \mathcal{A}=\mathrm{black\text{-}box}
\end{cases}
\right]$
\end{algorithmic}
}
\end{algorithm}

\subsubsection{Code Transformation Module}
This module generates semantically equivalent variants of a code sample via minimal, localized surface edits that preserve execution behavior and program structure. By altering lexical form while keeping semantics fixed, any systematic change in model behavior between the original and its variants is more likely attributable to memorization of the original lexical/syntactic patterns rather than semantic drift or major structural changes.

Given a code sample \( x \), this module generates a set of variants \( \mathcal{V} = \{x_1, x_2, \dots, x_n\} \). We focus on three distinct, localized transformation operations as follows:

\textbf{Function Name Renaming (\( \tau_f \))}: All function names (including method declarations and their call sites) are replaced with semantically equivalent alternatives. For instance, a function named \texttt{``solve''} might be renamed to \texttt{``compute''} or \texttt{``calculate''}. This transformation is applied consistently throughout the entire code snippet and any associated context.
    
\textbf{Variable Name Renaming (\( \tau_v \))}: All variable identifiers (local variables and parameters, etc.) are systematically replaced with alternative names. For example, a variable \texttt{``nums''} could be renamed to \texttt{``arr''} or \texttt{``numbers''}. Consistency is maintained across all references within the same scope.
    
\textbf{Test Case Transformation (\( \tau_t \))}: We use an LLM to propose new test inputs that are semantically similar to the originals, then run the original code to obtain outputs and verify consistency. Only regenerated tests that pass this execution-based validation replace the original ones, preserving semantic fidelity.

Formally, let the original sample be \( x_0 \). A transformed variant \( x_i \) is produced by applying a combination of these operations: \( x_i = (\tau_t \circ \tau_v \circ \tau_f)(x_0) \). By construction, the execution behavior of \( x_i \) remains identical to \( x_0 \), ensuring \( \mathcal{S}(x_0) = \mathcal{S}(x_i) \), where \( \mathcal{S} \) denotes the semantic functionality. Crucially, each transformation is designed to be minimal and localized, thereby maintaining a close lexical and structural resemblance to the original.

\subsubsection{Leakage Decision Module}
This module defines the detection score function $\mathcal{S}_M(\cdot)$ and the indicator function $\mathbb{I}$ of SrDetection under both gray-box and black-box settings. The specific computation and decision rules are described in the following sections.

\noindent\textbf{Gray-box Detection via Perplexity.}
Under the gray-box setting, we have access to the token-level probabilities of the target Code LLM \(M\). 
For a given set \(\mathcal{X} = \{x_0\} \cup \mathcal{V}\), consisting of the original sample \(x_0\) and its semantic variants \(\mathcal{V}\), the key observation is that, if the model has memorized \(x_0\), its perplexity will be lower than that of any variant. 

Accordingly, we define the detection score function \(\mathcal{S}_M(x)\) as:
\begin{equation}
    \mathcal{S}_M(x) \coloneqq \text{PPL}(x),
\end{equation}
where the perplexity of sample \(x\) is 
\(\text{PPL}(x) = \exp \big(- \frac{1}{|x|} \sum_{t=1}^{|x|} \log P_M(x_t \mid x_{<t}) \big)\), and \(x_t\) denotes the \(t\)-th token in the sequence. 

The final leakage decision under the gray-box setting is made using the indicator function \(\mathbb{I}[\cdot]\):
\[
\text{Leakage}_{\text{gray}}(x_0) = \mathbb{I}\Big[ \mathcal{S}_M(x_0) < \min_{x_i \in \mathcal{V}} \mathcal{S}_M(x_i) \Big],
\]
where \(\mathbb{I}[\cdot] = 1\) if the condition holds (indicating leakage) and 0 otherwise. 

By comparing scores only within the self-generated set \(\mathcal{X}\), this approach remains threshold-free and directly implements the self-referential decision principle of SrDetection.

\noindent\textbf{Black-box Detection via Generation Similarity.}
In the black-box setting, only the generated outputs of Code LLMs are accessible. 
For each sample \( x_i \in \mathcal{X} = \{x_0\} \cup \mathcal{V} \), we construct a prefix \( x_i^{\text{pre}} \) and feed it to the model \( M \) to obtain a generated suffix \( \hat{c}_i = M(x_i^{\text{pre}}) \). 
We denote the ground-truth suffix as \( c_i^{\text{gt}} \). 

Then, the detection score function under the black-box setting is defined as the n-gram overlap (NGO) between the generated and ground-truth suffix:
\begin{equation}
    \mathcal{S}_M(x_i) \coloneqq \text{NGO}(\hat{c}_i, c_i^{\text{gt}}),
\end{equation}
where the NGO of sample \(x\) is 
$\text{NGO}(\hat{c}_i, c_i^{\text{gt}}) = \frac{| \text{N-grams}(\hat{c}_i) \cap \text{N-grams}(c_i^{\text{gt}}) |}{| \text{N-grams}(c_i^{\text{gt}}) |}$, and \(\text{N-grams}(\cdot)\) denotes the set of all consecutive N-token sequences in a code snippet. 
This metric robustly captures the surface-level similarity of code sequences.

The final leakage decision under black-box setting is made using the indicator function \(\mathbb{I}[\cdot]\):
\[
\text{Leakage}_{\text{black}}(x_0) = \mathbb{I}\Big[ \mathcal{S}_M(x_0) > \max_{x_i \in \mathcal{V}} \mathcal{S}_M(x_i) \Big].
\]
Here, \(\mathbb{I}[\cdot]\) outputs 1 if the condition holds (indicating leakage) and 0 otherwise. 

Overall, by comparing detection scores only within the self-generated set \(\mathcal{X}\), this approach remains threshold-free 
and directly implements the self-referential decision principle of SrDetection.

\subsection{Leakage Testbed Construction}
To evaluate code leakage detection methods under precise conditions, we constructed a general testbed capable of supporting different datasets and open-source Code LLMs.

\noindent\textbf{Dataset and Model Preparation.}
We first selected a publicly available code benchmark dataset to serve as the evaluation corpus. To minimize pre-existing exposure, we filtered samples by computing the PPL of each original sample and its generated variants under the target model, removing any sample whose original form was already the easiest (lowest PPL) among its variants. The remaining samples were then split equally into training and test subsets to support controlled evaluation.

\noindent\textbf{Simulating Data Leakage via Continued Pre-Training.}
To emulate realistic code memorization, each model was further pre-trained on the training subset while keeping the test subset held out. To reflect typical pre-training pipelines, where code represents only a small fraction of the overall corpus, the training subset was mixed with a publicly available, well-known pre-training corpus. Detection performance was then evaluated on the held-out test subset under this mixed-corpus training setup.


\section{Experimental Settings}\label{sec:experimental_setup}
In this section, we present the evaluation metrics and a set of competitive baseline methods. Detailed implementation settings and dataset statistics are reported in Appendix~\ref{app:implementation_details}.

\begin{table*}[t]
\centering
\footnotesize
\setlength{\tabcolsep}{6pt} 
\renewcommand{\arraystretch}{1.08}
\caption{Detection performance comparison (F1-Macro, \%) between SrDetection and gray-box baselines on the continued pre-trained Qwen2.5-7B and Llama3.1-8B models across different training epochs. The \textbf{bold} number indicates the best performance, while the \underline{underlined} number represents the second-best.}
\label{tab:comparision-with-gray-box-baselines}
\begin{tabularx}{\textwidth}{@{} l l *{4}{r} *{4}{r} @{}}
\toprule
\textbf{Epochs} & \textbf{Methods} 
  & \multicolumn{4}{c}{\textbf{Qwen2.5-7B}} 
  & \multicolumn{4}{c}{\textbf{Llama3.1-8B}} \\
\cmidrule(lr){3-6} \cmidrule(lr){7-10}
 & & Acc. & Prec. & Rec. & F1 & Acc. & Prec. & Rec. & F1 \\
\midrule

\multirow{7}{*}{1}
 & PPL~\cite{Gonen0BSZ23}          & 50.85	&\underline{71.94}	&50.62	&34.88 & 50.34 & 64.03 & 50.11 & 33.77 \\
 & Lowercase~\cite{CarliniTWJHLRBS21}           & \underline{59.09}	&69.93	&\underline{58.92}	&\underline{52.44}& \underline{66.95} & 69.16 & \underline{66.88} & \underline{65.91} \\
 & Zlib~\cite{CarliniTWJHLRBS21}               & 50.24	&25.12	&50.00	&33.44 & 50.25 & 58.45 & 50.02 & 33.52 \\
 & Min-K\% Prob~\cite{ShiAXHLB0Z24}       & 50.43	&66.83	&50.19	&33.94& 50.57 & 68.06 & 50.34 & 34.31 \\
 & Min-K\%++ Prob~\cite{zhang2025mink}    & 54.31	&64.03	&54.11	&44.36 & 55.94 & 69.67 & 55.75 & 46.34 \\
 & DC-PDD~\cite{Zhang0GRFC24}            & 51.47&	70.75	&51.24	&36.36 & 53.54 & \underline{69.80} & 53.33 & 41.18 \\
 & \textbf{SrDetection} (Ours)                   & \textbf{74.95} & \textbf{75.00} & \textbf{75.81} & \textbf{74.76} & \textbf{70.01} & \textbf{70.10} & \textbf{70.03} & \textbf{69.99} \\
\midrule

\multirow{7}{*}{3}
 & PPL~\cite{Gonen0BSZ23}              & 50.38 & 75.15 & 50.14 & 33.75 & 50.39 & 75.15 & 50.16 & 33.79 \\
 & Lowercase~\cite{CarliniTWJHLRBS21}         & 52.27 & 73.51 & 52.05 & 37.98 & 53.94 & 75.78 & 53.73 & 41.26 \\
 & Zlib~\cite{CarliniTWJHLRBS21}               & 50.28 & 75.13 & 50.05 & 33.54 & 50.29 & 75.13 & 50.07 & 33.59 \\
 & Min-K\% Prob~\cite{ShiAXHLB0Z24}    & 50.90 & 75.29 & 50.67 & 34.90 & 50.93 & 75.29 & 50.71 & 34.98 \\
 & Min-K\%++ Prob~\cite{zhang2025mink}   & \underline{70.83} & \underline{80.62} & \underline{70.70} & \underline{68.22} & \underline{55.21} & \underline{76.21} & \underline{55.01} & \underline{43.72} \\
 & DC-PDD~\cite{Zhang0GRFC24}          & 54.88 & 75.84 & 54.66 & 43.10 & 52.40 & 74.67 & 52.19 & 38.18 \\
 & \textbf{SrDetection} (Ours)                     & \textbf{88.59} & \textbf{89.70} & \textbf{88.55} & \textbf{88.50} & \textbf{82.43} & \textbf{86.42} & \textbf{82.36} & \textbf{81.92} \\
\midrule

\multirow{7}{*}{5}
 & PPL~\cite{Gonen0BSZ23}         & 50.43 & 75.17 & 50.19 & 33.86 & 50.66 & 75.22 & 50.43 & 34.39 \\
 & Lowercase~\cite{CarliniTWJHLRBS21}        & 59.80 & 77.06 & 59.61 & 51.97 & 57.91 & 77.06 & 57.72 & 48.63 \\
 & Zlib~\cite{CarliniTWJHLRBS21}             & 50.28 & 75.13 & 50.05 & 33.54 & 50.27 & 75.12 & 50.05 & 33.54 \\
 & Min-K\% Prob~\cite{ShiAXHLB0Z24}  & 51.37 & 75.41 & 51.14 & 35.93 & 52.20 & 75.06 & 51.98 & 37.72 \\
 & Min-K\%++ Prob~\cite{zhang2025mink}  & \underline{78.55} & \underline{84.36} & \underline{78.45} & \underline{77.56} & \underline{60.29} & \underline{77.49} & \underline{60.11} & \underline{52.73} \\
 & DC-PDD~\cite{Zhang0GRFC24}        & 54.69 & 75.27 & 54.47 & 42.79 & 54.56 & 75.49 & 54.35 & 42.52 \\
 & \textbf{SrDetection} (Ours)                     & \textbf{90.53} & \textbf{91.56} & \textbf{90.49} & \textbf{90.47} & \textbf{84.47} & \textbf{87.96} & \textbf{84.41} & \textbf{84.09} \\
\bottomrule
\end{tabularx}
\end{table*}

\smallskip
\noindent\textbf{Evaluation Metrics.}
We evaluated detection performance using standard binary classification metrics: Accuracy, Precision, Recall, and F1-score.

\noindent\textbf{Baseline Methods.}
We compared SrDetection against competitive baselines in both gray-box and black-box settings:
\textbf{(1) Gray-box Baselines.}
These methods required access to token-level probabilities:
\textbf{PPL}~\cite{Gonen0BSZ23} used raw perplexity as the detection signal, where lower perplexity indicates a higher likelihood of memorization;
\textbf{Zlib}~\cite{CarliniTWJHLRBS21} computed the ratio between model perplexity and Zlib compression entropy to normalize for text complexity;
\textbf{Lowercase}~\cite{CarliniTWJHLRBS21} computed the perplexity ratio between the original text and its lowercase variant;
\textbf{Min-K\% Prob}~\cite{ShiAXHLB0Z24} used the average log probability of the lowest \(K\%\) tokens as the detection score, assuming non-member samples contain outlier tokens;
\textbf{Min-K\%++}~\cite{zhang2025mink} extended Min-K\% Prob with local-mode normalization for improved robustness;
\textbf{DC-PDD}~\cite{Zhang0GRFC24} incorporated divergence calibration between the model distribution and token frequency statistics.
\textbf{(2) Black-box Baselines.}
These methods operated using only model-generated text:
\textbf{VeilProbe}~\cite{VeilProbe} learned a sequence-to-sequence mapping between inputs and model outputs, using token perturbations to strengthen detection;
\textbf{DPDLLM}~\cite{zhou-etal-2024-dpdllm} used a reference model to extract features from generated text for membership classification.


\section{Results and Analysis}

\subsection{Main Results}


\begin{table*}[t]
\centering
\footnotesize
\setlength{\tabcolsep}{6pt}
\renewcommand{\arraystretch}{1.08}
\caption{Detection performance comparison (F1-Macro, \%) between our method and black-box baselines on the continued pre-trained Qwen2.5-7B and Llama3.1-8B models across different training epochs. The \textbf{bold} number indicates the best performance, while the \underline{underlined} number represents the second-best.}
\label{tab:comparision-with-black-box-baselines}
\begin{tabularx}{\textwidth}{@{} l l *{4}{r} *{4}{r} @{}}
\toprule
\textbf{Epochs} & \textbf{Methods} 
  & \multicolumn{4}{c}{\textbf{Qwen2.5-7B}} 
  & \multicolumn{4}{c}{\textbf{LLaMA3.1-8B}} \\
\cmidrule(lr){3-6} \cmidrule(lr){7-10}
 & & Acc. & Prec. & Rec. & F1 & Acc. & Prec. & Rec. & F1 \\
\midrule

\multirow{3}{*}{1}

 & DPDLLM~\cite{zhou-etal-2024-dpdllm}    & 49.62 & 49.42 & 11.97 & 41.22 & 50.23 & 50.23 & \textbf{100.00} & 33.59 \\
  & VeilProbe~\cite{VeilProbe} & \underline{51.04}	&\underline{51.10}	&\underline{50.96}	&\underline{49.33}	&\underline{51.56} & \underline{51.60}	&51.58	&\textbf{51.48} \\
 & \textbf{SrDetection} (Ours)  & \textbf{54.64} & \textbf{54.73} & \textbf{55.51} & \textbf{53.02} & \textbf{51.59} & \textbf{52.00} & \underline{51.68} & \underline{49.58} \\
\midrule

\multirow{3}{*}{3}

 & DPDLLM~\cite{zhou-etal-2024-dpdllm}    & 50.28 & 50.26 & \textbf{98.68} & 34.84 & 50.34 & 50.28 & \textbf{99.95} & 33.88 \\
  & VeilProbe~\cite{VeilProbe} &\underline{50.95}	&\underline{50.93}	&50.91	&\underline{50.68}	&\underline{51.27}&	\underline{51.28}&	51.28	&\underline{51.25} \\
 & \textbf{SrDetection} (Ours)  & \textbf{66.24} & \textbf{66.28} & \underline{66.78} & \textbf{66.00} & \textbf{63.06} & \textbf{63.45} & \underline{63.09} & \textbf{62.82} \\
\midrule
\multirow{3}{*}{5}

 & DPDLLM~\cite{zhou-etal-2024-dpdllm}    & 46.21 & 47.72 & \underline{73.89} & 41.70 & 49.07 & 49.64 & \textbf{95.35} & 34.99 \\
  & VeilProbe~\cite{VeilProbe} &\underline{51.75}	&\underline{51.83}	&51.69	&\underline{50.78}	&\underline{51.61}	&\underline{51.64}	&51.63	&\underline{51.53} \\
 & \textbf{SrDetection} (Ours)  & \textbf{81.91} & \textbf{81.89} & \textbf{82.27} & \textbf{81.85} & \textbf{78.60} & \textbf{79.02} & \underline{78.58} & \textbf{78.51} \\
\bottomrule
\end{tabularx}
\end{table*}

\begin{figure*}[ht]
    \centering
    \includegraphics[width=\textwidth]{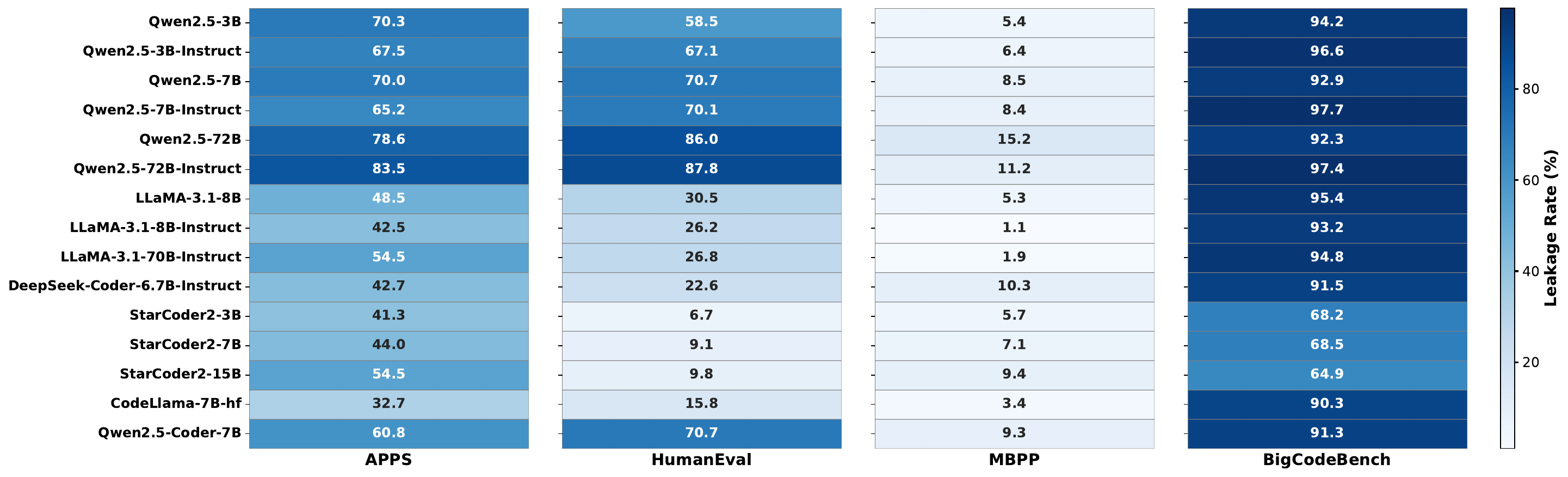}
    \caption{Data leakage rates (\%) of mainstream Code LLMs on standard benchmarks in the gray-box setting.}
    \label{fig:graybox_heatmap}
\end{figure*}


The performance of SrDetection against state-of-the-art gray-box and black-box baselines across different continued pre-training epochs is reported in Table~\ref{tab:comparision-with-gray-box-baselines} and Table~\ref{tab:comparision-with-black-box-baselines}. 
\smallskip

\noindent\textbf{Gray-box Setting.}
As shown in Table~\ref{tab:comparision-with-gray-box-baselines}, SrDetection consistently outperformed all gray-box baselines across both model families and all reported epochs. At 1 epoch, it achieved F1 scores of 74.76 on Qwen2.5-7B and 69.99 on LLaMA3.1-8B, substantially higher than competing approaches, many of which remained random predictions. Some baselines (e.g., Min-K\%++ and Lowercase) showed gains at specific epochs, but performance was unstable and model-dependent. In contrast, SrDetection remained strong and stable across epochs and models. This robustness arises from comparing each sample to its semantics-preserving variants, which provides an internal reference and avoids reliance on externally calibrated thresholds.

\noindent\textbf{Black-box Setting.}
Table~\ref{tab:comparision-with-black-box-baselines} reports black-box performance. On Qwen2.5-7B, SrDetection surpassed all baselines at every epoch, with F1 improving from 53.09 at 1 epoch to 82.06 at 5 epochs. On LLaMA3.1-8B, performance at 1 epoch was lower (F1 49.58) and comparable to VeilProbe. Nevertheless, SrDetection separated clearly from baselines at 3 and 5 epochs, indicating that the self-referential, code-aware comparison remained effective even under output-only access. Most baselines stayed near chance across epochs, underscoring the difficulty of black-box leakage detection and the advantage of relative comparison.

\noindent\textbf{Gray-box vs. Black-box.}
Overall, gray-box evaluation yielded higher absolute performance due to access to token probabilities, whereas black-box evaluation relied only on generated outputs. Importantly, both settings shared the same self-referential framework and semantics-preserving comparison, differing only in the evidence available to the detector. This consistency suggests that SrDetection’s effectiveness is driven by within-sample relative comparison rather than by specific access assumptions.

\begin{table}[t]
\centering
\caption{Detection performance (F1-Macro, \%) under different numbers of variants and training epochs under the gray-box setting. 
}
\label{tab:graybox_variants}
\footnotesize
\begin{tabular}{lcccc}
\toprule
\multirow{2}{*}{\textbf{Model}} & \multirow{2}{*}{\textbf{Variants ($n$)}} & \multicolumn{3}{c}{\textbf{Training Epochs}} \\
\cmidrule(lr){3-5}
 & & \textbf{1} & \textbf{3} & \textbf{5} \\
\midrule
\multirow{4}{*}{Llama-3.1-8B}
 & 1 & 33.43 & 33.43 & 33.43 \\
 & 3 & 65.30 & 69.65 & 72.03 \\
 & 5 & 67.20 & 74.89 & 77.05 \\
 & 10 & \textbf{69.99} & \textbf{81.92} & \textbf{84.09} \\
\midrule
\multirow{4}{*}{Qwen2.5-7B}
 & 1 & 33.44 & 33.44 & 33.44 \\
 & 3 & 69.32 & 75.56 & 77.45 \\
 & 5 & 71.56 & 81.44 & 83.05 \\
 & 10 & \textbf{74.76} & \textbf{88.50} & \textbf{90.47} \\
\bottomrule
\end{tabular}
\end{table}

\begin{figure}[ht]
    \centering
    \includegraphics[width=0.5\textwidth]{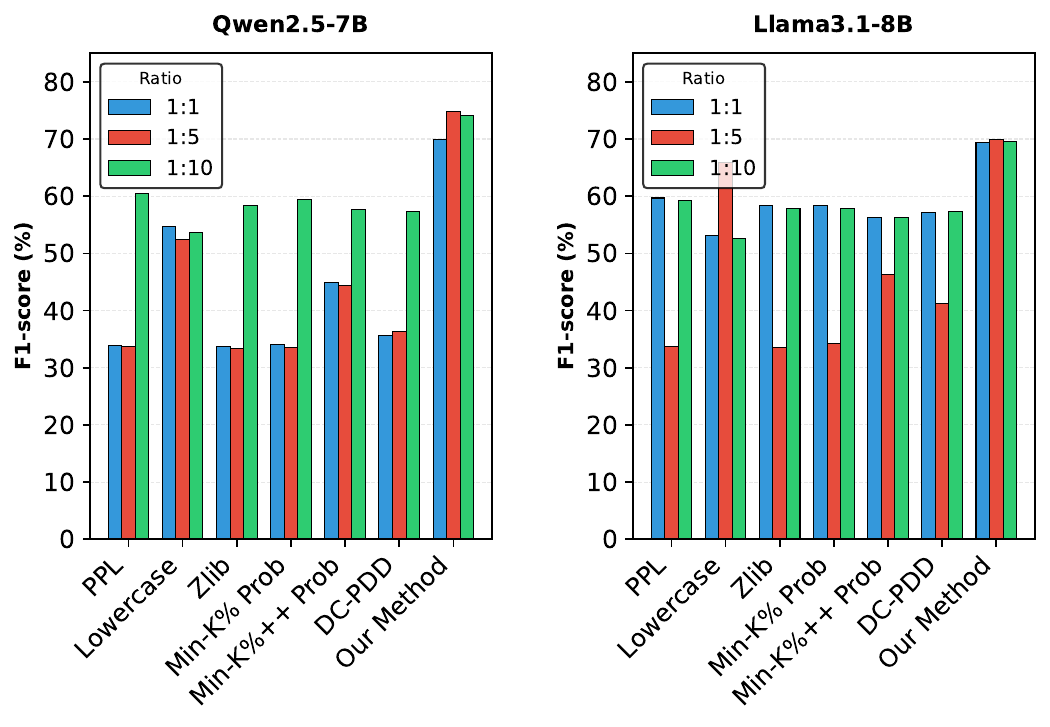
    }
    \caption{Detection performance (F1-score, \%) under different mixed ratios during continued pre-training with 1 epoch in gray-box setting.}
    \label{fig:max_ratio}
\end{figure}

\subsection{Sensitivity Analysis}
In the following, we examine the sensitivity of SrDetection under two key factors: (1) the number of semantics-preserving variants generated per original sample, and (2) the proportion of code in the continued pre-training corpus. 
Since gray-box evaluation provides more direct access to model logits and generally yields more stable performance, we focus our detailed analysis on the gray-box setting. Black-box sensitivity results, which rely solely on generated outputs, are reported in Appendix~\ref{sec:black_box_experiments}.

\noindent\textbf{Variants Quantity.} As shown in Table~\ref{tab:graybox_variants}, detection performance improved as the number of semantics-preserving variants increased, supporting the benefit of relative comparison between each original sample and its variants. Even with a small number of variants (e.g., \(n=3\)), SrDetection exceeded random baselines. Performance continued to improve over the tested range, suggesting that additional variants strengthen robustness across models and training epochs.

\noindent\textbf{Mixing Ratio.} 
Figure~\ref{fig:max_ratio} reports results under different code-to-text mixing ratios, obtained by mixing the training subset with a well-known, publicly available pre-training dataset~\cite{gao2020pile}. We evaluated ratios ranging from 1:1, 1:5, to 1:10 (code to general web data). Across all ratios, SrDetection consistently outperformed baseline methods. While probability-based baselines were sensitive to the data mix ratio, the self-referential, threshold-free design of SrDetection remained stable, demonstrating robustness to variations in pre-training corpus composition.

\subsection{Data Leakage across Benchmarks}
To assess leakage in widely used Code LLMs, we applied our gray-box detector to 15 publicly released models across four popular code generation benchmarks: APPS~\cite{HendrycksBKMAGB21}, HumanEval~\cite{Chen21Humaneval}, MBPP~\cite{Jacob25mbpp}, and BigCodeBench~\cite{ZhuoVCH0WYZHPB025}, as summarized in Figure~\ref{fig:graybox_heatmap}. Leakage was evaluated directly from model behavior under gray-box access.

APPS exhibited consistently higher leakage for several Qwen2.5 variants, in line with overlap-based trends reported in prior work~\cite{LessLeak-Bench}, which used MinHash+LSH near-duplicate detection followed by manual verification. HumanEval showed moderate leakage, consistent with previous analyses indicating small but nonzero contamination. MBPP was largely unaffected for most models, reflecting its low leakage in the same study. Notably, BigCodeBench demonstrated high leakage in our gray-box evaluation, suggesting that behavior-based detection can reveal potential leakage beyond strict data overlap.

Overall, these results largely align with overlap-based findings while providing complementary, benchmark-specific insights. Related black-box experiments are reported in Appendix~\ref{sec:black_box_experiments}.
\begin{table}[t]
\centering
\footnotesize
\caption{Thresholds and scores of baselines on sample \#1061 under the gray-box setting. All baseline methods failed to detect leakage because their scores did not fall below the respective thresholds, whereas our method correctly identified the leaked sample.}

\label{tab:baseline-results}
\begin{tabular}{@{}lccc@{}}
\toprule
\textbf{Method} & \textbf{Score} & \textbf{Threshold} &\textbf{Correct?} \\
\midrule
PPL & -1.9485 & -21.6516 & \ding{55} \\
Lowercase & -0.9486 & -0.9820  & \ding{55} \\
Zlib & -0.0008 & -0.0195  & \ding{55} \\
Min-K\% & -2.7820 & -7.9785  & \ding{55} \\
Min-K\%++ & -0.6970 & -1.5204  & \ding{55} \\
DC-PDD & 0.0098 & 0.0090  & \ding{55} \\
Our Method & -- & -- & \checkmark \\
\bottomrule
\end{tabular}
\end{table}

\lstset{
  style=borderedcode,
  basicstyle=\ttfamily\scriptsize,
  tabsize=2,
  columns=fullflexible,
  basewidth=0.5em,
  keepspaces=true,
  lineskip=-1pt,
  aboveskip=2pt,
  belowskip=2pt
}

\begin{figure}[t]
\centering

\begin{minipage}{0.48\linewidth}
\begin{lstlisting}

def shipWithinDays(weights, D):
  left = max(weights)
  right = left * len(weights) // D
  while left < right:
    mid = left + (right - left) // 2
    c = 0
    d = 1
    for w in weights:
      if c + w <= mid:
        c += w
      else:
        d += 1
        c = w
    if d > D:
      left = mid + 1
    else:
      right = mid
  return left
# ...
\end{lstlisting}
\centerline{(a) PPL=1.71}
\end{minipage}
\hfill
\begin{minipage}{0.48\linewidth}
\begin{lstlisting}

def getMinimumDaysForShipping(cargo_weights, num_days):
  left_limit = max(cargo_weights)
  right_limit = left_limit * len(cargo_weights) // num_days
  while left_limit < right_limit:
    mid_range = left_limit + (right_limit - left_limit) // 2
    weight_sum = 0
    load_count = 1
    for cargo in cargo_weights:
      if weight_sum + cargo <= mid_range:
        weight_sum += cargo
      else:
# ... 
\end{lstlisting}
\centerline{(b) PPL=2.31\vspace{1pt}}
\end{minipage}

\caption{Comparison of an original code sample (a) and one of its semantics-preserving variants (b).}
\label{fig:compact-code-parallel}
\end{figure}

\subsection{Case Study}
To illustrate why the self-referential design of the proposed SrDetection method succeeds where conventional detectors fail, we analyze some cases and present a representative code sample. As shown in Table~\ref{tab:baseline-results}, established gray-box baselines, including PPL, Zlib, and Min-K\%, failed to detect leakage for this example. These methods typically rely on absolute scores and fixed thresholds; here, all scores remained above their thresholds, producing a false negative and underscoring the brittleness of thresholded, natural-language-oriented detectors on code.

In contrast, SrDetection evaluates each sample \emph{relative} to its semantics-preserving variants. For the sample in Figure~\ref{fig:compact-code-parallel}, we generated ten variants that preserved core logic while altering surface form (e.g., variable and function names). The original sample yielded the minimum perplexity (PPL=1.7082), while all variants had higher perplexity (2.0287--2.7565). This pronounced local minimum indicates memorization tied to the original surface form. As shown in Figure~\ref{fig:compact-code-parallel}, which compares the original sample with one variant, SrDetection correctly flagged leakage while threshold-based baselines failed.


\section{Discussion and Conclusion}
SrDetection reframes data leakage detection for Code LLMs as a \emph{self-referential} problem: instead of depending on inaccessible pretraining corpora, fragile temporal rules, or externally calibrated thresholds, it measures leakage through \emph{relative} behavioral discrepancies between an original code sample and its semantics-preserving variants. This perspective is particularly well matched to code benchmarks, where surface forms can be altered without changing functionality, and where reuse and refactoring routinely break timestamp- and overlap-based assumptions. 
To support rigorous evaluation, we further introduce a controlled testbed that enables precise attribution of leakage by explicitly separating seen and unseen samples through continued pre-training. This testbed is model- and dataset-agnostic, and allows leakage detectors to be compared under reproducible conditions.
Empirically, our results show that SrDetection can reliably identify leaked samples in both gray-box and black-box settings and can surface benchmark-specific contamination patterns that are not revealed by overlap checks alone. Beyond detection performance, these findings suggest that leakage is not a marginal artifact but a systematic confounder in current evaluation practice, reinforcing the need for leakage-aware benchmark design and reporting.

\section*{Limitations}
While SrDetection demonstrates robust performance in detecting memorization of code samples, there remain opportunities for further improvement. The effectiveness of detection depends on the diversity and quality of semantics-preserving variants; exploring additional transformation strategies could strengthen the contrastive signal. In black-box settings, subtle output differences may be masked in very large or instruction-tuned models, suggesting that alternative or complementary signals could further enhance detection. Future work may also extend SrDetection to cover broader code modalities and to integrate adaptive transformation strategies that better reflect real-world code variability.

\section*{Acknowledgments}
We would like to thank the anonymous reviewers for their valuable comments and suggestions to improve this paper.
We used AI assistants only for minor language polishing and grammar correction. All research ideas, methodology, experiments, and conclusions were developed and verified by the authors. Min Yang was supported by National Key Research and Development Program of China (2024YFF0908200), National Natural Science Foundation of China (Grant No. 62376262), the Natural Science Foundation of Guangdong Province of China (2024A1515030166, 2025B1515020032). Shiwen Ni was supported by GuangDong Basic and Applied Basic Research Foundation (2023A1515110718 and 2024A1515012003) and Shenzhen Science and Technology Program  (JCYJ20250604182917023).
\bibliography{custom}
\clearpage
\appendix

\section{Implementation Details}\label{app:implementation_details}
\subsection{SrDetection}
\begin{table}[h]
\centering
\caption{Statistics of the filtered APPS-based datasets after PPL-based exclusion.}
\footnotesize
\label{appendix:dataset_statistics}
\setlength{\tabcolsep}{6pt}
\begin{tabular}{c cc}
\toprule
\multirow{2}{*}{\textbf{Model}} 
& \multicolumn{2}{c}{\textbf{Number of Instances}} \\
\cmidrule(lr){2-3}
& \textbf{Training} & \textbf{Test} \\
\midrule
Qwen2.5-7B     & 1061 & 1051 \\
LLaMA-3.1-8B   & 2216 & 2196 \\
\bottomrule
\end{tabular}
\end{table}
\begin{table}[h] 
\centering 
\footnotesize 
\setlength{\tabcolsep}{4pt} 
\renewcommand{\arraystretch}{1.1} 
\caption{Threshold values used by baseline methods across training epochs (1, 3, 5) for Qwen2.5-7B and LLaMA3.1-8B, organized by method and epoch.}
\label{tab:thresholds_by_method} 
\begin{tabular}{lcc} 
\toprule 
\textbf{Method} & \textbf{Qwen2.5-7B} & \textbf{LLaMA3.1-8B} \\ 
\midrule
\multicolumn{3}{c}{\textbf{1 Epoch}} \\
PPL             & -21.6516 & -23.4101 \\
Lowercase       & -0.9820  & -0.9681  \\
Zlib            & -0.0195  & -0.0188  \\
Min-K\% Prob    & -7.9785  & -7.9442  \\
Min-K\%++ Prob  & -1.5204  & -1.5833  \\
DC-PDD          & 0.0090   & 0.0091   \\
\midrule
\multicolumn{3}{c}{\textbf{3 Epochs}} \\
PPL             & -36.1829 & -36.1829 \\
Lowercase       & -1.0078  & -1.0078  \\
Zlib            & -0.0217  & -0.0217  \\
Min-K\% Prob    & -9.5799  & -9.5799  \\
Min-K\%++ Prob  & -5.8413  & -5.8413  \\
DC-PDD          & 0.0082   & 0.0082   \\
\midrule
\multicolumn{3}{c}{\textbf{5 Epochs}} \\
PPL             & -86.3298 & -86.3298 \\
Lowercase       & -1.0021  & -1.0021  \\
Zlib            & -0.0272  & -0.0272  \\
Min-K\% Prob    & -11.8104 & -11.8104 \\
Min-K\%++ Prob  & -18.4400 & -18.4400 \\
DC-PDD          & 0.0073   & 0.0073   \\
\bottomrule 
\end{tabular} 
\end{table}

All experiments are implemented using PyTorch\footnote{\url{https://pytorch.org}}
 and the Hugging Face Transformers library\footnote{\url{https://huggingface.co/docs/transformers}}
, running on a system with four NVIDIA A100 GPUs. We evaluate our proposed method, SrDetection, under both gray-box and black-box scenarios. For SrDetection, we generate $n=10$ semantic-preserving variants per code sample to enable self-referential comparison. In the black-box setting, we additionally employ N-gram overlap (NGO) with $N=7$.
To create semantic variants, we design prompt templates targeting two key transformations: identifier renaming (Table~\ref{tab:prompt_identifier}) and test case replacement (Table~\ref{tab:prompt_testcase}). These templates preserve the core logic, functionality, and overall code structure of the original samples while introducing sufficient syntactic diversity to challenge conventional detectors. 
Table~\ref{tab:thresholds_by_method} provides a supplementary reference by reporting the threshold values used by the baseline methods across different models with different training epochs (1, 3, 5), corresponding to the performance results shown in Table~\ref{tab:comparision-with-gray-box-baselines} in the main text.

\begin{table}[h]
\caption{Prompt template for identifier renaming.}
\label{tab:prompt_identifier}
\centering
\footnotesize
\begin{tabular}{|p{7.5cm}|}
\hline
\textbf{User:}\\
You are given a Python code snippet:
\\[2pt]
\textit{\{\{python\_code\_snippet\}\}}
\\[6pt]
\textbf{Task:}\\
Generate 10 alternative versions of the snippet by renaming function names, parameters, and variable identifiers.
\\[6pt]
\textbf{Constraints:}\\
-- Preserve comments, formatting, indentation, spacing, and line breaks exactly.\\
-- Update identifiers consistently if they appear in comments.\\
-- Do not change the logic or functionality.
\\[6pt]
Return each version as a single string, using \textbackslash n for newlines.\\
\hline
\end{tabular}
\end{table}

\begin{table}[h]
\caption{Prompt template for test case transformation.}
\label{tab:prompt_testcase}
\centering
\footnotesize
\begin{tabular}{|p{7.5cm}|}
\hline
\textbf{User:}\\
You are given a programming problem:
\\[2pt]
\textit{\{\{question\}\}}
\\[6pt]
\textbf{Provided Test Cases:}\\
\textbf{Input:} \textit{\{\{test\_case\_input\}\}}\\
\textbf{Output:} \textit{\{\{test\_case\_output\}\}}
\\[6pt]
\textbf{Task:}\\
Replace the sample test cases in the problem with the provided examples.
\\[6pt]
\textbf{Constraints:}\\
-- Preserve structure, formatting, indentation, and line breaks.\\
-- Only replace input/output sections.\\
-- Do not modify any other text.
\\[6pt]
Return the modified problem as a single string, using \textbackslash n for newlines.\\
\hline
\end{tabular}
\end{table}

\subsection{Leakage Testbed}
In our study, we instantiated the leakage testbed using the APPS dataset~\cite{HendrycksBKMAGB21} and selected representative open-source Code LLMs as target models, including Qwen2.5-7B~\cite{Qwen2.5} and LLaMA-3.1-8B~\cite{Llama-3.1}. Table~\ref{appendix:dataset_statistics} reports the number of instances in the filtered APPS-based datasets after PPL-based exclusion. To support controlled evaluation, the dataset was split into training and test subsets at a 1:1 ratio, resulting in model-specific dataset sizes summarized in Table~\ref{appendix:dataset_statistics}.
To simulate realistic pre-training conditions in which code constitutes a minor portion of the overall corpus, the APPS training subset was mixed with general web data from RefinedWeb~\cite{gao2020pile}. A mixing ratio of 1:5 (code to general web data) was adopted to approximate typical code concentrations. All models undergo continued pre-training on our filtered datasets using LoRA~\cite{HuSWALWWC22} with a learning rate of $1.0 \times 10^{-4}$, batch size of 32, and training epochs of 1, 3, and 5.

\section{Supplementary Black-box Experiments}\label{sec:black_box_experiments}

In addition to the gray-box analyses reported in the main text, we conducted a series of experiments under the black-box setting. 

\textbf{Black-box Detection Across Models and Benchmarks.}
Table~\ref{tab:model-results-basic-single} presents data leakage detection rates (\%) for several widely used Code LLMs across three benchmarks: HumanEval, MBPP, and BigCodeBench. Our method achieves consistently higher detection rates compared with baseline approaches, highlighting its ability to identify potential leakage even with limited access to model internals. Notably, leakage is more pronounced on BigCodeBench for most models, consistent with trends observed in gray-box evaluations.

\textbf{Sensitivity to Data Composition.}
Table~\ref{tab:mixed_ratios_compact} reports F1-Macro performance under varying code-to-text mixing ratios during continued pre-training (1 epoch). SrDetection maintains strong, stable performance across all ratios (1:1, 1:5, 1:10), while baseline methods exhibit more variable results, demonstrating the resilience of our self-referential, threshold-free approach under different data distributions.

\textbf{Impact of Variant Quantity.}
Table~\ref{tab:blackbox_variants} shows performance under varying numbers of semantic-preserving variants. Increasing the number of variants generally improves detection, confirming that relative comparisons between the original sample and its variants are effective even when only the generated outputs are available. This trend mirrors observations in the gray-box setting, indicating that the self-referential mechanism is robust to the level of access provided.

Overall, these black-box experiments complement the gray-box analyses, demonstrating that SrDetection’s relative, variant-based framework can reliably detect code leakage even under restricted observational settings.

\begin{table}[h]
\centering
\footnotesize
\setlength{\tabcolsep}{4pt} 
\renewcommand{\arraystretch}{1.1} 
\caption{Data leakage detection rates (\%) across different Code LLMs and benchmarks (black-box).}
\label{tab:model-results-basic-single}
\begin{tabular}{p{2.5cm}ccc}
\toprule
\textbf{Model} & \textbf{Humaneval} & \textbf{MBPP} & \textbf{BigCodeBench} \\
\midrule
Qwen2.5-7B                   & 16.46 & 10.68 & 52.94 \\
LLaMA-3.1-8B                 & 7.93  & 9.45  & 61.19 \\
DeepSeek-Coder-6.7B-Instruct & 7.93  & 10.78 & 35.21 \\
StarCoder2-7B                 & 7.93  & 6.16  & 35.03 \\
CodeLlama-7B-hf               & 11.59 & 12.73 & 49.96 \\
Qwen2.5-Coder-7B              & 18.90 & 18.17 & 44.78 \\
\bottomrule
\end{tabular}
\end{table}
\begin{table}[ht] 
\centering 
\footnotesize 
\setlength{\tabcolsep}{3pt} 
\caption{Detection performance (F1-Macro, \%) under different mixed ratios during continued pre-training with 1 epoch under the black-box setting.}
\label{tab:mixed_ratios_compact} 
\begin{tabular}{@{}llcc@{}} 
\toprule 
\textbf{Mixed Ratio} & \textbf{Method} & \textbf{Qwen2.5-7B} & \textbf{Llama3.1-8B} \\
\midrule \multirow{7}{*}{1:1} & PPL & 33.86 & 59.68 \\ & Lowercase & 54.68 & 53.11 \\ 
& Zlib & 33.65 & 58.45 \\ 
& Min-K\% Prob & 34.03 & 58.44 \\ & Min-K\%++ Prob & 44.85 & 56.28 \\ 
& DC-PDD & 35.69 & 57.12 \\ 
& \textbf{SrDetection} (Ours)  & \textbf{69.86} & \textbf{69.33} \\ 
\midrule \multirow{7}{*}{1:5} & PPL & 33.75 & 33.77 \\ 
& Lowercase & 52.44 & \underline{65.91} \\ 
& Zlib & 33.44 & 33.52 \\ 
& Min-K\% Prob & 33.61 & 34.31 \\ 
& Min-K\%++ Prob & 44.36 & 46.34 \\ 
& DC-PDD & 36.36 & 41.18 \\ & \textbf{SrDetection} (Ours)  & \textbf{74.76} & \textbf{69.99} \\ 
\midrule \multirow{7}{*}{1:10} & PPL & 60.46 & 59.29 \\ & Lowercase & 53.68 & 52.58 \\ 
& Zlib & 58.32 & 57.82 \\ 
& Min-K\% Prob & 59.41 & 57.86 \\ 
& Min-K\%++ Prob & 57.70 & 56.27 \\ & DC-PDD & 57.24 & 57.24 \\ 
& \textbf{SrDetection} (Ours)  & \textbf{74.13} & \textbf{69.58} \\ \bottomrule 
\end{tabular} 
\end{table} 

\begin{table}[h] 
\centering 
\caption{Detection performance (F1-Macro, \%) under different numbers of variants and training epochs under the black-box setting. } 
\label{tab:blackbox_variants} 
\footnotesize \begin{tabular}{lcccc} \toprule
\multirow{2}{*}{\textbf{Model}} & \multirow{2}{*}{\textbf{Variants ($n$)}} & \multicolumn{3}{c}{\textbf{Training Epochs}} \\
\cmidrule(lr){3-5} & & \textbf{1} & \textbf{3} & \textbf{5} \\
\midrule \multirow{4}{*}{Qwen2.5-7B} 
& 1 & 52.57 & 58.35 & 65.17 \\
& 3 & 55.15 & 63.39 & 75.67 \\ 
& 5 & 54.61 & 64.86 & 78.67 \\
& 10 & 53.02 & 66.00 & 81.85 \\ 
\midrule 
\multirow{4}{*}{Llama-3.1-8B}
& 1 & 50.96 & 55.91 & 64.45 \\
& 3 & 51.95 & 60.05 & 72.38 \\
& 5 & 51.48 & 61.74 & 74.74 \\
& 10 & 49.58 & 62.82 & 78.51 \\ 
\bottomrule
\end{tabular}
\end{table}

\end{document}